# ADVERSARIAL SYNTHESIS LEARNING ENABLES SEGMENTATION WITHOUT TARGET MODALITY GROUND TRUTH

*Yuankai Huo[1], Zhoubing Xu[1], Shunxing Bao[2], Albert Assad[3],*
*Richard G. Abramson[4] and Bennett A. Landman[1,2,4]*

[1] Electrical Engineering, Vanderbilt University, Nashville, TN, USA 37235
[2] Computer Science, Vanderbilt University, Nashville, TN, USA 37235
[3] Incyte Corporation, Wilmington, DE, USA 19803
[4] Radiology and Radiological Science, Vanderbilt University, Nashville, TN, USA 37235

## ABSTRACT

A lack of generalizability is one key limitation of deep learning based segmentation. Typically, one manually labels new training images when segmenting organs in different imaging modalities or segmenting abnormal organs from distinct disease cohorts. The manual efforts can be alleviated if one is able to reuse manual labels from one modality (e.g., MRI) to train a segmentation network for a new modality (e.g., CT). Previously, two stage methods have been proposed to use cycle generative adversarial networks (CycleGAN) to synthesize training images for a target modality. Then, these efforts trained a segmentation network independently using synthetic images. However, these two independent stages did not use the complementary information between synthesis and segmentation. Herein, we proposed a novel end-to-end synthesis and segmentation network (EssNet) to achieve the unpaired MRI to CT image synthesis and CT splenomegaly segmentation simultaneously without using manual labels on CT. The end-to-end EssNet achieved significantly higher median Dice similarity coefficient (0.9188) than the two stages strategy (0.8801), and even higher than canonical multi-atlas segmentation (0.9125) and ResNet method (0.9107), which used the CT manual labels.

*Index Terms*— splenomegaly, segmentation, unpaired, multi-model, synthesis

## 1. INTRODUCTION

Splenomegaly, the condition of having an abnormally large spleen (e.g., >500 cubic centimeter), is a biomarker for liver disease, infection and cancer. Previous automated methods have been proposed to perform segmentation on normal spleens [1, 2] and with splenomegaly [3-5]. Recently, deep convolutional neural network (DCNN) based methods have been used in splenomegaly and shown superior performance [6, 7]. However, one major limitation of deploying DCNN methods is that one typically has to manually trace a new set of training data when segmenting organs in a new imaging modality or segmenting abnormal organs from a new disease cohort. For instance, a DCNN trained with normal spleens was not able to capture the spatial variations of splenomegaly (**Fig. 1**). Therefore, a straightforward solution is to manually annotate a set of splenomegaly CT scans. However, manual tracing is resource intensive and potentially error prone.

Image synthesis has been used to segment images for one modality from another modality [8-11]. However, paired images were typically required for traditional synthesis methods. Recently, the cycle generative adversarial networks (CycleGAN) [12] image synthesis methods provided an effective tool for inter-modality synthesis from unpaired images [13, 14]. Therefore, one could synthesize the training images and labels for splenomegaly patients and labels on one modality (e.g., MRI) while targeting another image modality (e.g., CT). Upon such idea, Chartsias et al. [15] proposed an CT to MRI synthesis method using CycleGAN and trained another independent MRI segmentation network (called "Seg.") using the synthesized MRI images. Although still using manual labels for both modalities, this two stage framework (called "CycleGAN+Seg.") revealed a promising direction: segmentation was possible without ground truth in the target modality.

In this paper, we propose a novel end-to-end synthesis and segmentation network (EssNet) to perform MRI to CT synthesis and CT splenomegaly segmentation simultaneously without using ground truth labels in CT. The EssNet was trained by unpaired MRI and CT scans and only used manual labels from MRI scans.

## 2. DATA

Unpaired 60 whole abdomen MRI T2w scans and 19 whole abdomen CT with splenomegaly spleen were used as the experimental data, whose imaging parameters and demographic information were introduced in [6] and [3]. Six labels (spleen, left kidney, right kidney, liver, stomach and body) were manually delineated for each MRI [6], while one label (spleen) was manually traced for each CT scan [3]. Additional 75 whole abdomen CT scans with normal spleens [2] were used to train a baseline DCNN method.

## 3. METHOD

The network structure of EssNet is shown in **Fig. 2,** where "*A*" indicates MR images while "*B*" represents CT images. The 9 block ResNet (defined in [12, 16]) was used as the two generators ($G_1$ and $G_2$). $G_1$ synthesized an image $x$ in modality $A$ to the generated $B$ image ($G_1(x)$), while $G_2$ synthesize an image $y$ in modality $B$ to the generated $A$ image ($G_2(y)$). The PatchGAN (defined in [12, 17]) was employed as the two adversarial discriminators ($D_1$ and $D_2$). $D_1$ distinguished if the CT image was real or generated, while $D_2$ determined for the MR image. When deploying such framework on unpaired $A$ and $B$, two training paths (Path A and Path B) existed in forward cycles. The cycle synthesis subnet was basically the same as CycleGAN [12].

Since the aim of the proposed EssNet was to perform end-to-end synthesis and segmentation. The segmentation network $S$ was concatenated after $G_1$ directly as an additional forward branch in Path A. The 9 block ResNet [12, 16] were used as $S$, whose network structure was identical to $G_1$. Then, the estimated segmentation from generated B was derived.

Five loss functions were used to train the network. Two adversarial loss functions $\mathcal{L}_{\text{GAN}}$ were defined as

$$\mathcal{L}_{\text{GAN}}(G_1, D_1, A, B) = E_{y \sim B}[\log D_1(y)] \\ + E_{x \sim A}[\log(1 - D_1(G_1(x)))] \quad (1)$$

$$\mathcal{L}_{\text{GAN}}(G_2, D_2, B, A) = E_{x \sim A}[\log D_2(x)] \\ + E_{y \sim B}[\log(1 - D_2(G_2(y)))] \quad (2)$$

Two cycle consistency loss $\mathcal{L}_{\text{cycle}}$ functions were used to compare the reconstructed images with real images.

$$\mathcal{L}_{\text{cycle}}(G_1, G_2, A) = E_{x \sim A}[\|G_2(G_1(x)) - x\|_1] \quad (3)$$

$$\mathcal{L}_{\text{cycle}}(G_2, G_1, B) = E_{y \sim B}[\|G_1(G_2(y)) - y\|_1] \quad (4)$$

The segmentation loss function was defined as

$$\mathcal{L}_{\text{seg}}(S, G_1, A) = - \sum_i m_i \cdot \log(S(G_1(x_i))) \quad (5)$$

where $m$ was the manual labels for image $x$, $i$ was the index of a pixel. Then, the total loss function was defined as

$$\mathcal{L}_{\text{total}} = \lambda_1 \cdot L_{\text{GAN}}(G_1, D_1, A, B) + \lambda_2 \cdot \mathcal{L}_{\text{GAN}}(G_2, D_2, B, A) \\ + \lambda_3 \cdot \mathcal{L}_{\text{cycle}}(G_1, G_2, A) + \lambda_4 \cdot \mathcal{L}_{\text{cyle}}(G_2, G_1, B) \\ + \lambda_5 \cdot \mathcal{L}_{\text{seg}}(S, G_1, A) \quad (6)$$

In this work, the lambdas were empirically set to $\lambda_1 = 1$, $\lambda_2 = 1$, $\lambda_3 = 10$, $\lambda_4 = 10$, $\lambda_5 = 1$. To minimize the $\mathcal{L}_{\text{total}}$, the Adam optimizer was used [12]. The examples of real, synthesized, reconstructed and segmentation images for Path A and Path B were shown in **Fig. 3**.

In testing, only trained network S was used and $B'$ represented the testing CT images. the Dice similarity coefficient (DSC) values between automated and manual segmentations were used as the metrics to evaluate the performance of different segmentation methods. All statistical significance tests were made using a Wilcoxon signed rank test (p<0.05).

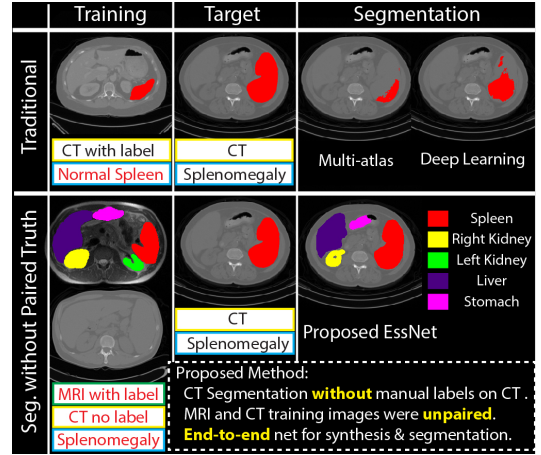

**Fig. 1.** The upper row shown that carnonical methods trained by normal spleen failed in splenomegaly segmentation. The lower row shown that the proposed EssNet was able to achieve splenomegaly segmentation from unpaired MRI and CT training images without using CT labels.

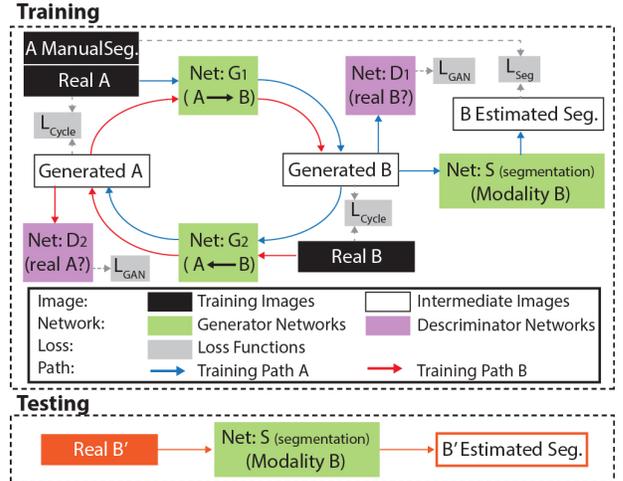

**Fig. 2.** The left side was the CycleGAN synthesis subnet, where $A$ was MRI and $B$ was CT. $G_1$ and $G_2$ were the generators while $D_1$ and $D_2$ were discriminators. The right subnet was the segmentation subnet for an end-to-end training. Loss function were added to optimize the EssNet.

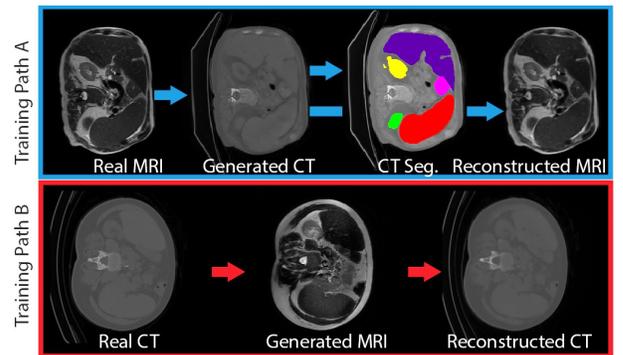

**Fig. 3.** The qualitative results of the synthesized images and segmentations in training Path A and Path B.

## 4. EXPERIMENTS

### 4.1. CT Segmentation with CT Manual Labels

Our previous Spleen Segmentation Network (SSNet) [7] was trained by 75 CT scans with normal spleens to see if such network was able to segment 19 splenomegaly CT scans. Then, two baseline methods were employed using 19 splenomegaly CT scans in both training and testing. The first baseline approach was the adaptive Gaussian mixture model multi-atlas segmentation (AGMM MAS) [3]. The leave-one-subject-out validation strategy was used to evaluate AGMM MAS method using the same parameter as [3]. The second approach was to train a particular 9 block ResNet [12, 16] for each subject using leave-one-subject-out strategy on 19 CT splenomegaly scans. ResNet had the same network structure and hyperparameters as the net $S$ in EssNet. The hyperparameters of $S$ are introduced in the next section.

### 4.2. CT Segmentation without CT Manual Labels

The two stage CycleGAN+Seg. method from Chartsias et al. [15] and the proposed EssNet were evaluated. The network structure of CycleGAN and Seg. were the same as the cycle subnet and segmentation subnet in **Fig. 2** to enable a fair comparison between independent and end-to-end frameworks. The training data, testing data and network hyperparameters of two methods were also the same. Briefly, 60 splenomegaly MRI scans and 19 splenomegaly CT scans were used to train the EssNet. The manually traced spleen labels on 60 MRI scans were used in training. The manually traced spleen labels on 19 splenomegaly CT scans were only used for the evaluation purpose and were not been used in any training procedures in CycleGAN+Seg. or EssNet.

The axial view slices for all scans were resampled to 256x256 resolution. In total, 3262 MRI slices and 1874 CT slices were used in the experiments. The number of input and output channels of all networks are all one except S, which had seven output channels. The learning rate of Adam was 0.0001 for $G_1$, $G_2$ and $S$ and 0.0002 for $D_1$ and $D_2$. All networks were trained and validated from 1 to 100 epochs. The epoch with highest mean DSC between predicted and manual segmentation on 19 splenomegaly CT scans were reported in the results. The best performance of ResNet (epoch=90) was obtained from leave-one-subject-out internal validation. The best performance of SSNet (epoch=10), CycleGAN+Seg. (epoch=50) and EssNet (epoch=40) were evaluated from the external validation since labels for 19 splenomegaly CT scans were never used in the training.

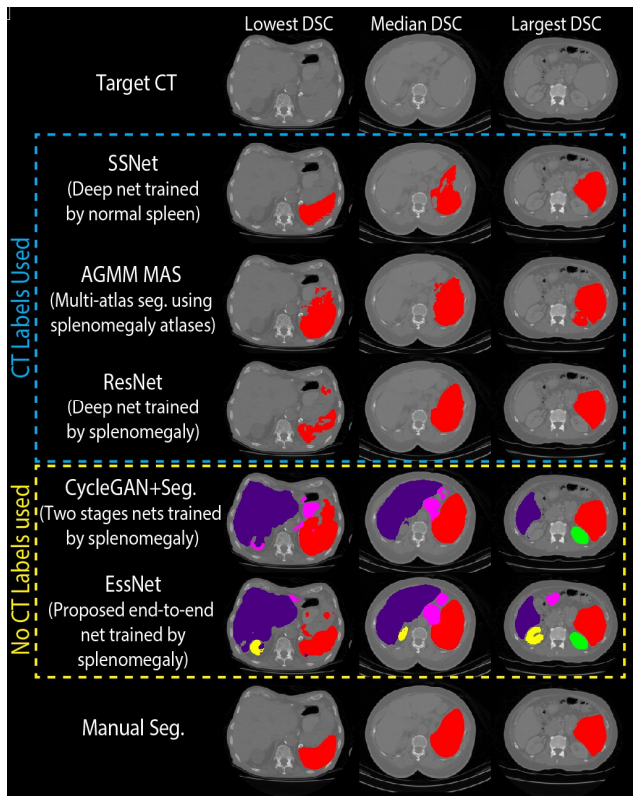

**Fig. 4.** The qualitative results of (1) three canonical methods using CT manual labels in CT segmentation, and (2) CycleGAN+Seg. and the proposed EssNet methods without using CT manual labels. The splenomegaly CT labels were only used in validation and excluded from training for (2). Moreover, later methods were not only able to perform spleen segmentation but also estimated labels for other organs, which were not provided by canonical methods when such labels were not available on CT.

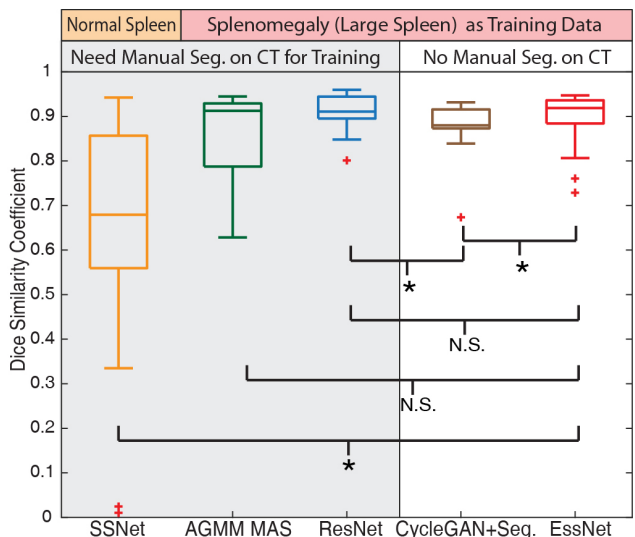

**Fig. 5.** The boxplot results of all CT splenomegaly testing images, where "*" means the difference are significant at p<0.05, while "N.S." means not significant.

## 5. RESULTS

The qualitative results of different methods on three subjects (lowest, median and highest DSC for the EssNet) were shown in the **Fig. 4**. From the results, the EssNet was not only able to perform the spleen segmentation, but also estimated segmentations on liver, left kidney, right kidney and stomach.

The quantitative results of different segmentation strategies on all CT scans were shown in the **Fig. 5** as a boxplot. The "*" indicates the difference between methods were significant, while "N.S." means not significant.

## 6. CONCLUSION AND DISCUSSION

In this work, we proposed the end-to-end EssNet for simultaneous image synthesis and segmentation. We demonstrate this approach on splenomegaly CT segmentation without using ground truth labels in CT. From **Fig. 3**, the proposed end-to-end approach was able to achieve MRI to CT synthesis, CT to MRI synthesis, and the CT segmentation simultaneously. **Fig. 4** shown that the proposed method was not only able to obtain spleen segmentation but also estimate liver, kidney, stomach labels, while the canonical methods using CT data only were not able to when such labels were not available on CT. **Fig. 5** shown that the SSNet trained by normal spleen CT images was significantly worse than other methods. The proposed EssNet method was significantly better than the two stages CycleGAN+Seg. method. Without using CT labels, the EssNet achieved the comparable performance as the AGMM MAS and ResNet that used CT labels. On the contrary, the performance of CycleGAN+Seg. was significantly worse than ResNet.

This study opens the possibility of using EssNet to perform the segmentations on other modalities on which target labels are not known and paired inter-modality data are not available. An interesting limitation of this work is that the networks are 2-D (but assessed in 3-D) due to time and memory concerns. Either post processing for 3-D consistency or 3D EssNet would be a promising area for development.

## 7. ACKNOWLEDGEMENT

This research was supported by NSF CAREER 1452485, NIH 5R21EY024036, 1R21NS064534, 2R01EB006136, 1R03EB012461, R01NS095291, and InCyte Corporation (Abramson/Landman). It was support from Intramural Research Program, National Institute on Aging, NIH and in part using the resources of the Advanced Computing Center for Research and Education (ACCRE) at Vanderbilt University. It was supported by ViSE/VICTR VR3029 and the National Center for Research Resources, Grant UL1 RR024975-01, and is now at the National Center for Advancing Translational Sciences, Grant 2 UL1 TR000445-06. We appreciate NIH S10 Shared Instrumentation Grant 1S10OD020154-01, Vanderbilt IDEAS grant, and ACCRE's Big Data TIPs grant from Vanderbilt University.